\documentclass[a4paper, 10 pt, conference]{ieeeconf}  
\IEEEoverridecommandlockouts                              
\usepackage{amsmath,amsfonts}
\usepackage{url}
\usepackage{adjustbox}
\usepackage{algorithm}
\usepackage{algpseudocode}
\usepackage{amsmath}
\usepackage{pifont}
\usepackage{booktabs}
\usepackage{tabularx}
\usepackage{multirow}
\usepackage{tabularx}

\usepackage{graphicx}
\usepackage{subcaption} 
\usepackage{float}
\usepackage{threeparttable}
\usepackage{array}
\usepackage{multirow}
\usepackage{array}

\usepackage{xltabular}
\usepackage{notoccite}
\usepackage[noadjust]{cite}
\usepackage{stfloats}

\usepackage{cuted}
\usepackage{caption}

\begin{document}

\title{D3DWA: Adaptive Weight and Prediction-Horizon for Dynamic Window Approach via Dueling Double Deep Q-Network}
\author{
Zahra Jooyandeh$^{\dag 1}$, Masato Kobayashi$^{\dag 1,2*}$, Yuki Uranishi$^{1}$
\thanks{
$^{\dag}$ Equal Contribution, 
$^{1}$ The University of Osaka, 
$^{2}$ Kobe University,
$^*$ corresponding author: kobayashi.masato.cmc@osaka-u.ac.jp
}
}

\maketitle
\begin{abstract}
The Dynamic Window Approach (DWA) is widely used for local navigation, but its performance depends strongly on parameters that are typically fixed before navigation. In particular, the appropriate prediction horizon can vary with local free space: longer horizons support efficient motion in open areas, whereas shorter horizons help preserve feasible motions in narrow or cluttered regions. This paper proposes D3DWA, an adaptive DWA framework based on a Dueling Double Deep Q-Network (D3QN), which jointly selects the DWA evaluation weights and prediction horizon from a continuous navigation state at every control step while retaining DWA’s trajectory generation and collision checking. In eight simulated environments, including unseen layouts, D3DWA reached every goal.
Real-robot experiments further showed that D3DWA completed all three tested configurations, including a constrained case in which the weights-only variant timed out. These results demonstrate the benefit of jointly adapting the evaluation weights and prediction horizon.
Additional material is available at \url{https://mertcookimg.github.io/d3dwa/}.
\end{abstract}

\section{Introduction}

Autonomous mobile robots are increasingly used for indoor service tasks such as
delivery, cleaning, guidance, and inventory management. During a single
navigation task, a robot may encounter open spaces, narrow corridors, doorways,
and densely cluttered areas, each requiring different navigation behavior.
Because such environmental conditions can change along the same route, a local
planner should not only generate feasible motions in real time but also adapt
its behavior to the current situation.

The Dynamic Window Approach (DWA)~\cite{fox1997dwa} is a widely used local
path-planning method that samples dynamically reachable velocity commands and
evaluates their predicted trajectories according to criteria such as goal
progress, obstacle clearance, and speed. These criteria are combined using
predefined evaluation weights, while the prediction horizon determines how far
each candidate trajectory is projected. A short horizon can provide more
reactive behavior in narrow or cluttered areas, whereas a long horizon can
support efficient motion in open spaces. However, both the evaluation weights
and the prediction horizon are typically fixed before navigation.

Adaptive approaches have therefore been developed to tune DWA parameters
online. In particular, DQDWA~\cite{10237217} uses tabular Q-learning to select
evaluation weights according to the surrounding situation, improving the
flexibility of DWA without replacing the underlying planner. However, its
navigation state must be manually discretized into a limited number of states,
which restricts the use of richer observations. Moreover, DQDWA adapts only the
evaluation weights while keeping the prediction horizon fixed.

\begin{figure}[t]
  \centering
  \includegraphics[width=1.0\columnwidth]{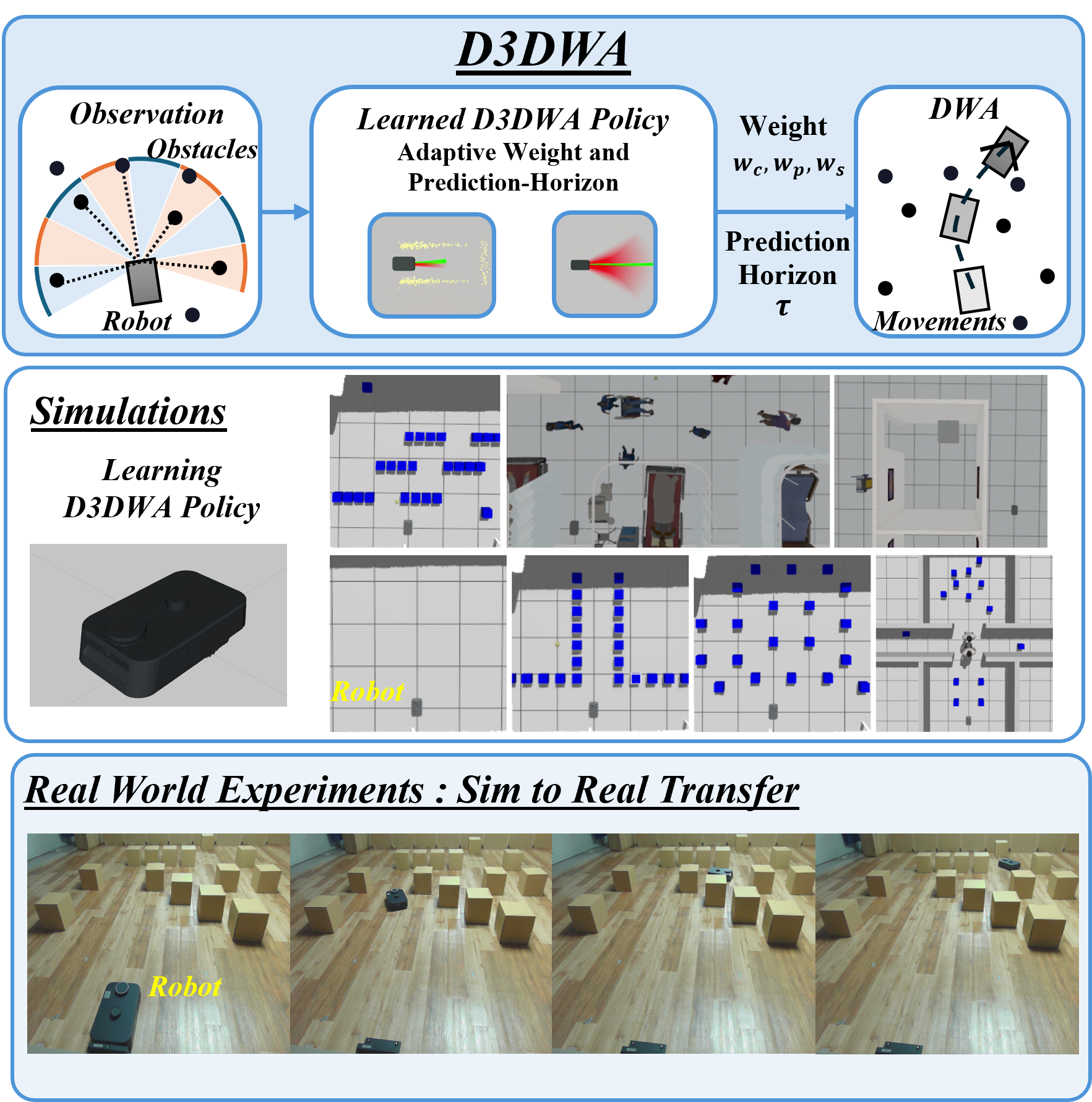}
  \caption{Concept of D3DWA.}
  \label{fig:usecase}
\end{figure}

To address these limitations, we propose D3DWA, an adaptive DWA method based on
a Dueling Double Deep Q-Network (D3QN), as illustrated in
Fig.~\ref{fig:usecase}. D3DWA estimates action values directly from a continuous
navigation state and jointly selects the DWA evaluation weights and prediction
horizon at every control step. The DWA planner itself remains responsible for
trajectory generation, collision checking, and velocity-command selection.
The prediction horizon is a complementary adaptation dimension: it determines
which portion of the future a candidate trajectory is scored over, and
therefore provides a degree of freedom that the evaluation weights alone
cannot substitute for.
We compare D3DWA with fixed-parameter DWA, DQDWA, and a D3QN-based variant that
adapts only the evaluation weights, thereby isolating the contribution of
prediction-horizon adaptation. The complete system is evaluated in multiple
simulated environments and on a real mobile robot.

The main contributions of this study are as follows:
\begin{itemize}
    \item We extend adaptive DWA from a tabular, discretized-state formulation
    to a D3QN-based formulation that operates on a continuous navigation state.
    \item We jointly adapt the DWA evaluation weights and prediction horizon and
    evaluate the contribution of horizon adaptation through a weight-only
    ablation.
    \item We implement the proposed method in a ROS~2 navigation system and
    validate it against fixed-parameter DWA and DQDWA in both simulation and
    real-robot experiments.
\end{itemize}

\begin{figure*}[t]
  \centering
  \includegraphics[width=\textwidth]{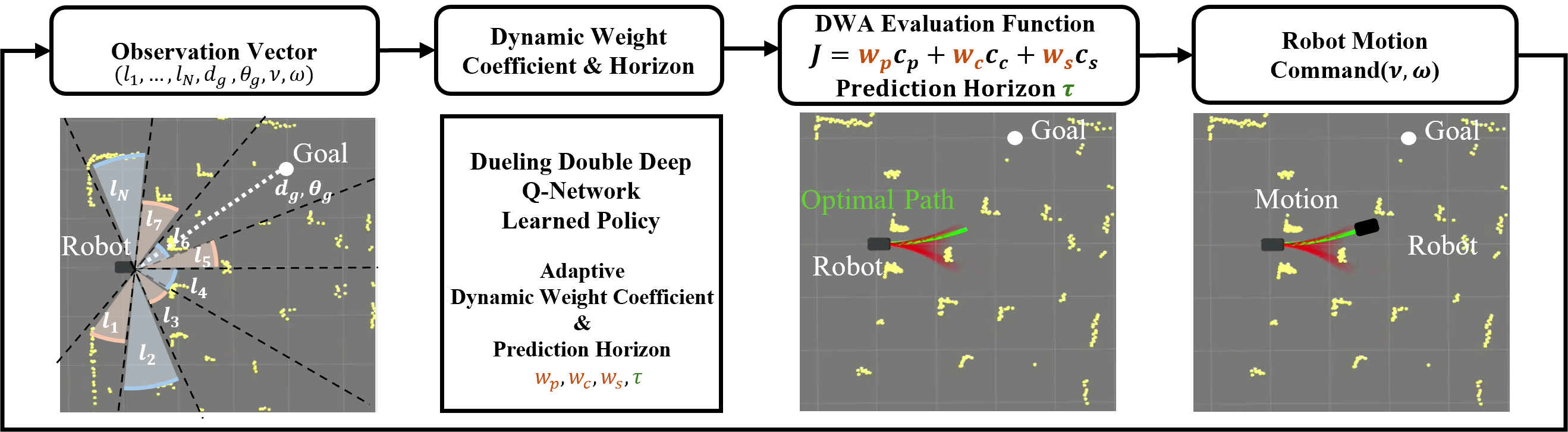}
  \caption{\textbf{System Overview of D3DWA}}
  \label{fig:system}
\end{figure*}

\section{Related Work}

\subsection{Reinforcement Learning for Navigation}

Deep reinforcement learning has been widely applied to robot navigation,
including end-to-end approaches that directly map sensor observations to
velocity commands~\cite{tai2017virtual,chen2017socially}. Although these methods
can adapt to complex environments, they often require substantial training data,
can be difficult to stabilize, and may suffer from sim-to-real transfer
issues~\cite{dulacarnold2021challenges,zhao2020sim2real,xiao2022survey}.
An alternative is to retain a classical local planner and use learning only to
adapt its parameters, thereby preserving explicit trajectory generation and
collision checking.

Value-based reinforcement learning is well suited for selecting planner
parameters from a discrete set. However, tabular Q-learning~\cite{watkins1992qlearning}
, requires predefined state discretization and scales poorly as the state
representation becomes richer. Deep Q-Networks (DQN)~\cite{mnih2015dqn} address
this limitation through the function approximation. Double DQN~\cite{vanhasselt2016double} reduces overestimation in value estimation,
while dueling networks~\cite{wang2016dueling} separately estimate state value
and action advantage. In this work, we combine these techniques as a Double
Dueling Double Deep Q-Network (D3QN) to select DWA parameters from continuous
navigation observations.

\subsection{Adaptive Parameter Tuning for DWA}

The Dynamic Window Approach (DWA)~\cite{fox1997dwa} is a widely used local
planner that evaluates dynamically reachable trajectories according to weighted
objectives such as goal progress, obstacle clearance, and speed. DWA has been
extended in terms of navigation-system integration, convergence, global-plan
guidance, and trajectory optimization
~\cite{macenski2020nav2,ogren2005convergent,zhang2019globaldwa,rosmann2017teb}.
However, its evaluation weights are generally selected offline and remain fixed
during navigation.

Several studies have therefore investigated the online adaptation of planner
parameters. Fuzzy-logic approaches adjust DWA weights according to environmental
conditions~\cite{abubakr2022fuzzy}. The Adaptive Planner Parameter Learning
framework~\cite{xiao2022appl} learns to tune classical planner parameters from
demonstrations~\cite{xiao2020appld}, corrective interventions
~\cite{wang2021appli}, or reinforcement learning~\cite{xu2021applr}.
Within the DWA family, DQDWA~\cite{10237217}, which is most closely related to
our work, uses tabular Q-learning to select evaluation weights from a discretized
navigation state. More recent approaches employ neural networks or actor--critic
methods to adapt DWA weights from richer observations
~\cite{dobrevski2020adaptive,dobrevski2024dadwa,martini2024social}.

Most of these approaches adapt the evaluation weights while keeping the prediction horizon fixed. Other studies modify trajectory generation within a fixed horizon. Liu et al.~\cite{liu2024td3dwa} and Jiang et al.~\cite{jiang2024agv} use TD3 to improve trajectory sampling beyond conventional constant-velocity DWA, but the prediction horizon remains fixed. In contrast, our method treats the prediction horizon as part of the action and jointly selects it with the evaluation weights at every control step using D3QN.

\begin{figure*}[t]
  \centering
  \begin{subfigure}[t]{1.0\textwidth}
    \includegraphics[width=\linewidth]{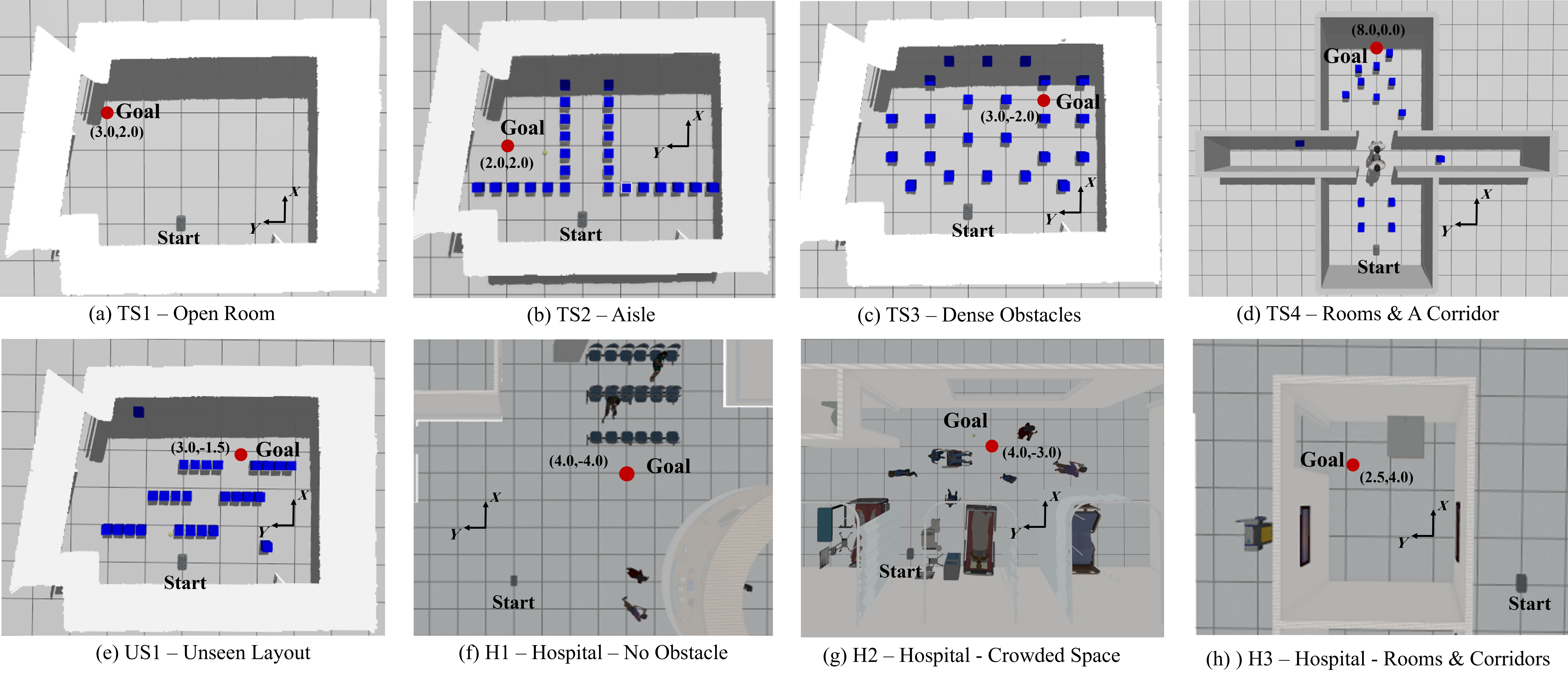}
  \end{subfigure}\hfill

  \caption{\textbf{Evaluation Environments.}}
  \label{fig:scenes}
\end{figure*}
\section{Method}
\subsection{System Overview}
At every D3DWA control step a body-frame
observation is built from the lidar scan and wheel odometry; a D3QN agent maps it
to a discrete action; the action decodes to a DWA parameter tuple and is published
to the planner; and the DWA planner uses that tuple to score and execute a velocity command. Fig.~\ref{fig:system} shows the discussed architecture.

\subsection{DWA Execution Backbone}
DWA forms the dynamic window from the current velocity and the acceleration
limits, and draws a grid of $N_v \times N_\omega$ samples $(v,\omega)$ from it. 
Each sample is rolled forward under a constant-velocity model over a prediction
horizon $\tau$, and the predicted trajectory is scored by three criteria:
progress $c_p$, the reduction in goal distance between the start and the end of
the trajectory; clearance $c_c$, the smallest distance from the trajectory to a
scan point; and speed $c_s = |v|$. Progress is therefore measured as a distance
reduction rather than by the heading term of the original
formulation~\cite{fox1997dwa}.
The three criteria are then
min--max normalised to $[0,1]$ \emph{across the admissible set}, and the executed
command maximises
\begin{equation}
  J(v,\omega) = w_{p}\,\hat{c}_{p} + w_{c}\,\hat{c}_{c} + w_{s}\,\hat{c}_{s},
  \label{eq:dwa_objective}
\end{equation}
where $\hat{c}_{p},\hat{c}_{c},\hat{c}_{s}$ are the normalised criteria and
$(w_{p},w_{c},w_{s})$ the weights supplied by the agent. 

\subsection{State Space}

The lidar readings are partitioned into $K$ equal angular sectors, each holding
the minimum range among its beams normalised by a maximum usable range $d_{\max}$. The state space is the concatenation of this
summary with the goal geometry and the robot's motion,
\begin{equation}
  s = \big[\, d_g,\; \theta_g,\; v,\; \omega,\; l_1, \dots, l_K \,\big] \in \mathbb{R}^{K+4},
  \label{eq:obs}
\end{equation}
where $d_g$ is the goal distance clipped at a horizon $d_g^{\max}$ and normalised,
$\theta_g$ the body-frame goal bearing normalised to $[-1,1]$, and $v,\omega$ the measured velocities normalised by their limits. Because the sector summary is already a hand-engineered feature rather than a raw scan, no convolutional backbone is required and the state stays small enough to evaluate inside the control loop.

\subsection{Action Space}
\label{sec:actions}
The agent selects the three weights and the prediction horizon simultaneously, as a single discrete action:
\begin{equation}
  a \mapsto (w_p, w_c, w_s, \tau), \quad
  w_\bullet \in \mathcal{W}, \;\; \tau \in \mathcal{T},
  \label{eq:action}
\end{equation}
for a finite weight candidate set $\mathcal{W}$ and horizon set $\mathcal{T}$. The
Cartesian product contains $|\mathcal{W}|^3|\mathcal{T}|$ tuples. Since only the
weight ratios matter, the uniform triples $(c,c,c)$ are equivalent and only the
smallest is retained, which for the candidate set used here removes every
duplicate:
\begin{equation}
  |\mathcal{A}| = \big(|\mathcal{W}|^3 - |\mathcal{W}| + 1\big)\,|\mathcal{T}| .
  \label{eq:nactions}
\end{equation}
Restricting $\mathcal{T}$ to a single value recovers a weights-only action space, which serves both as the ablation isolating the horizon's contribution and as the action space of the tabular baseline, so the two are matched by construction.

Selecting the pair jointly rather than with two heads is deliberate, because the horizon and weights interact: a short horizon with a high clearance weight avoids
nearby obstacles without spending lookahead on segments that will be replanned
before they are reached, whereas a long horizon with a high progress weight lets the robot commit to direct paths once space opens up.

\begin{table*}[t]
\centering
\caption{Simulation results following the Case~S1 protocol of~\cite{10237217}.
TL and PD denote trajectory length and movement posture displacement, respectively.
\textsf{T} denotes timeout. Best values in each environment and metric are shown in bold.}
\label{tab:main}

\scriptsize
\setlength{\tabcolsep}{2.0pt}
\renewcommand{\arraystretch}{1.10}

\begin{tabular}{l*{8}{ccc}}
\toprule
\multirow{2}{*}{Method}
& \multicolumn{3}{c}{TS1}
& \multicolumn{3}{c}{TS2}
& \multicolumn{3}{c}{TS3}
& \multicolumn{3}{c}{TS4}
& \multicolumn{3}{c}{US1}
& \multicolumn{3}{c}{H1}
& \multicolumn{3}{c}{H2}
& \multicolumn{3}{c}{H3} \\
\cmidrule(lr){2-4}
\cmidrule(lr){5-7}
\cmidrule(lr){8-10}
\cmidrule(lr){11-13}
\cmidrule(lr){14-16}
\cmidrule(lr){17-19}
\cmidrule(lr){20-22}
\cmidrule(lr){23-25}

& Time & TL & PD
& Time & TL & PD
& Time & TL & PD
& Time & TL & PD
& Time & TL & PD
& Time & TL & PD
& Time & TL & PD
& Time & TL & PD \\
\midrule

DWA I $(1,1,2)$
& 18.53 & \textbf{3.58} & 0.93
& \textsf{T} & -- & --
& \textsf{T} & -- & --
& \textsf{T} & -- & --
& \textsf{T} & -- & --
& 22.11 & 7.30 & 5.25
& \textsf{T} & -- & --
& \textsf{T} & -- & -- \\

DWA II $(1,2,1)$
& 40.42 & 6.90 & 14.76
& 29.65 & 7.09 & \textbf{3.61}
& 29.20 & 4.51 & 6.41
& \textsf{T} & -- & --
& \textsf{T} & -- & --
& 51.19 & 8.04 & 5.23
& 24.32 & 6.84 & 4.07
& \textsf{T} & -- & -- \\

DWA III $(2,1,1)$
& 17.08 & \textbf{3.58} & 0.90
& \textsf{T} & -- & --
& 29.48 & 4.40 & 5.58
& \textsf{T} & -- & --
& 47.63 & 5.29 & 7.22
& 24.45 & \textbf{6.01} & \textbf{1.79}
& 22.03 & \textbf{5.13} & \textbf{1.73}
& 51.15 & 11.83 & 13.42 \\

DQDWA~\cite{10237217}
& \textbf{16.59} & \textbf{3.58} & 0.96
& \textsf{T} & -- & --
& 49.43 & 9.54 & 17.23
& \textsf{T} & -- & --
& 46.09 & 5.38 & \textbf{6.95}
& 58.41 & 9.86 & 13.53
& \textbf{21.56} & 5.14 & 1.81
& \textsf{T} & -- & -- \\

D3DWA (w/o horizon)
& 18.05 & 3.59 & 0.92
& \textsf{T} & -- & --
& 28.56 & 4.42 & 5.90
& \textsf{T} & -- & --
& 37.39 & 5.81 & 7.35
& 36.45 & 10.04 & 10.26
& 25.26 & 6.67 & 4.70
& 37.82 & \textbf{9.95} & 6.26 \\

\textbf{D3DWA (proposed)}
& 17.01 & 3.59 & \textbf{0.88}
& \textbf{29.22} & \textbf{7.02} & 3.62
& \textbf{24.11} & \textbf{4.01} & \textbf{4.46}
& \textbf{44.12} & \textbf{9.73} & \textbf{7.72}
& \textbf{35.73} & \textbf{4.93} & 9.46
& \textbf{19.27} & 6.26 & 2.81
& 25.85 & 5.64 & 7.53
& \textbf{32.84} & 10.04 & \textbf{5.95} \\

\bottomrule
\end{tabular}
\end{table*}

\subsection{D3QN Network}
The action-value function is approximated by a dueling network: a shared multilayer perceptron over the observation feeds
two heads, a scalar state-value stream $V$ and an advantage stream $A$ with one
output per action, recombined as
\begin{equation}
  Q(s,a;\theta) = V(s;\theta) + A(s,a;\theta) - \tfrac{1}{|\mathcal{A}|}\textstyle\sum_{a'} A(s,a';\theta).
  \label{eq:dueling}
\end{equation}
The decomposition is well suited to this action space because most actions differ
from their neighbours in a single weight or horizon value, so $V$ can be estimated
from every transition irrespective of which action was taken.

Learning follows the Double DQN formulation, in which the next action is chosen by
the online network and evaluated by a target network $\theta^{-}$:
\begin{equation}
  y = r + \gamma\,(1-d)\; Q\!\big(s', \arg\max_{a'} Q(s',a';\theta);\, \theta^{-}\big),
  \label{eq:double}
\end{equation}
with discount $\gamma$ and terminal indicator $d$. Decoupling selection from
evaluation curbs the max operator's overestimation bias, which grows with the size
of the action set and so matters more here than for the tabular baseline.
Transitions are stored in a replay buffer and sampled uniformly, and the loss is
the Huber error between $Q(s,a;\theta)$ and $y$. The target network is
soft-updated at every step, and exploration is $\varepsilon$-greedy with an
exponentially decaying schedule advanced only by training steps.

\subsection{Reward}
The reward follows the structure of the baseline~\cite{10237217}, scaled uniformly
by 1/100. Positive scaling leaves the optimal policy unchanged, so
the comparison remains fair, while keeping the Bellman targets in a range the
function approximator can track. The reward $R$ is
defined as
\begin{equation}
  R = R_1 + R_2 + R_3 .
  \label{eq:reward}
\end{equation}
$R_1$ is a reward related to the
episode outcome; goal, collision or timeout:
\begin{equation}
  R_1 =
  \begin{cases}
    r_{\mathrm{goal}}                        & \text{if reach goal}\\
    r_{\mathrm{coll}} + r_{\mathrm{step}}  & \text{else if collide or time out}\\
    r_{\mathrm{step}}                        & \text{otherwise}
  \end{cases}
  \label{eq:r1}
\end{equation}
$R_2$ is a reward related to the distance from the goal position:
\begin{equation}
  R_2 =
  \begin{cases}
     r_{\mathrm{gc}}  & \text{if get close to goal position}\\
     -r_{\mathrm{gc}} & \text{else if farther from goal position}\\
    0                         & \text{otherwise}
  \end{cases}
  \label{eq:r2}
\end{equation}
$R_3$ is a reward related to the distance from the nearest obstacle:
\begin{equation}
  R_3 =
  \begin{cases}
    -r_{\mathrm{oc}} & \text{if approach obstacle}\\
    r_{\mathrm{oc}}  & \text{else if go away from obstacle}\\
    0                         & \text{otherwise}
  \end{cases}
  \label{eq:r3}
\end{equation}
The magnitudes $r_{\mathrm{goal}}$, $r_{\mathrm{coll}}$, $r_{\mathrm{step}}$,
$r_{\mathrm{gc}}$ and $r_{\mathrm{oc}}$ are those of the baseline and are listed in Table~\ref{tab:params}. 

\section{Simulation}

\subsection{Simulation Setup}
\label{sec:sim-conditions}

The simulation system was implemented using the Robot Operating System (ROS)
and Gazebo. The parameters used for the planner and reinforcement-learning
agent are summarized in Table~\ref{tab:params}.

Six methods were compared. Three fixed-parameter DWA configurations were used
following the baseline setting: DWA~I
$\{w_p,w_c,w_s\}=\{1,1,2\}$, DWA~II $\{1,2,1\}$, and DWA~III
$\{2,1,1\}$, all with a fixed prediction horizon of $\tau=4$\,s.
DQDWA~\cite{10237217} was used as the tabular Q-learning baseline.
D3DWA (w/o horizon) uses the same D3QN formulation as the proposed method but
adapts only the evaluation weights while keeping $\tau=4$\,s. The full D3DWA
jointly adapts the evaluation weights and prediction horizon.

Following~\cite{10237217}, navigation performance was evaluated using the time
to reach the goal, trajectory length (TL), and movement posture displacement
(PD). PD represents the accumulated change in robot heading during navigation.

\begin{table}[t]
\centering
\caption{Parameter values instantiating the framework of Section~III.}
\label{tab:params}
\scriptsize
\setlength{\tabcolsep}{3pt}
\begin{tabular}{llc}
\toprule
Group & Parameter & Value\\
\midrule
\multirow{7}{*}{Platform}
 & lidar field of view                       & $240^\circ$\\
 & robot radius $r$ / safety margin $m$      & $0.15$ / $0.05$\,m\\
 & max.\ translational velocity              & $0.4$\,m/s\\
 & max.\ angular velocity                    & $2.0$\,rad/s\\
 & max.\ translational acceleration          & $2.0\,m/s^2$\\
 & max.\ angular acceleration                & $3.0\,rad/s^2$\\
 & control period $\Delta t$                 & $0.1$\,s (10\,Hz)\\
\midrule
\multirow{4}{*}{Observation}
 & sectors $K$                               & $8$ ($30^\circ$ each)\\
 & max.\ usable range $d_{\max}$             & $3.5$\,m\\
 & goal-distance clip $d_g^{\max}$           & $6.0$\,m\\
 & observation dimension $K\!+\!4$           & $12$\\
\midrule
\multirow{3}{*}{Action}
 & weight candidates $\mathcal{W}$           & $\{1,2,3\}$\\
 & horizon candidates $\mathcal{T}$          & $\{2,3,4,5\}$\,s\\
 & actions $|\mathcal{A}|$, Eq.~(\ref{eq:nactions})  & $100$\\
\midrule
\multirow{2}{*}{Planner}
 & velocity samples $N_v \times N_\omega$    & $20 \times 40$\\
 & rollout resolution                        & $0.1$\,s\\
\midrule
\multirow{3}{*}{Termination}
 & goal tolerance $d_{\mathrm{goal}}$        & $0.10$\,m\\
 & collision threshold $d_{\mathrm{coll}}$   & $0.10$\,m\\
 & episode time limit $T_{\max}$             & $60$\,s\\
\midrule
\multirow{3}{*}{Reward}
 & goal / collision / timeout                & $+50$ / $-2$ / $-2$\\
 & per-step penalty                          & $-0.02$\\
 & goal / obstacle shaping                   & $\pm0.1$ / $\pm0.05$\\
\bottomrule
\end{tabular}
\end{table}

\begin{figure*}[t]
  \centering
  \includegraphics[width=\textwidth]{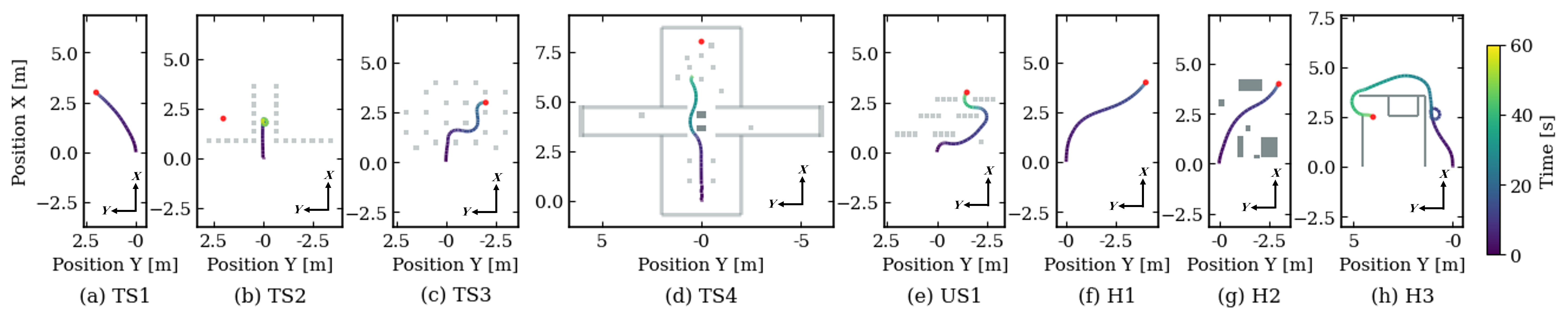}
  \caption{Executed trajectories of DWA III
  ($\{w_p, w_c, w_s\}=\{2,1,1\}$), coloured by elapsed time.
  The robot starts at the origin and the goal is marked in red.}
  \label{fig:traj-dwa3}
\end{figure*}

\begin{figure*}[t]
  \centering
  \includegraphics[width=\textwidth]{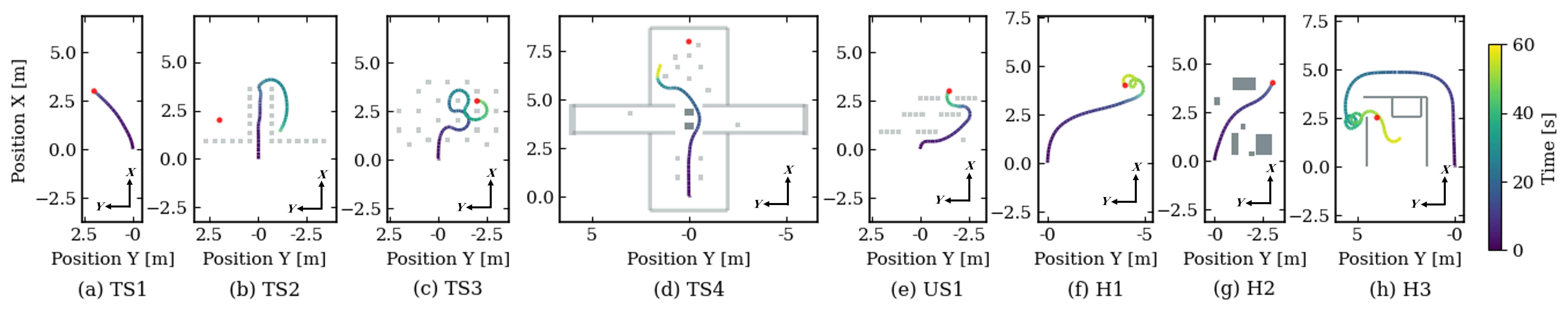}
  \caption{Executed trajectories of the tabular baseline
  DQDWA~\cite{10237217}, coloured by elapsed time.
  The robot starts at the origin and the goal is marked in red.}
  \label{fig:traj-dqdwa}
\end{figure*}
\begin{figure*}[t]
    \centering
    \includegraphics[width=\textwidth]{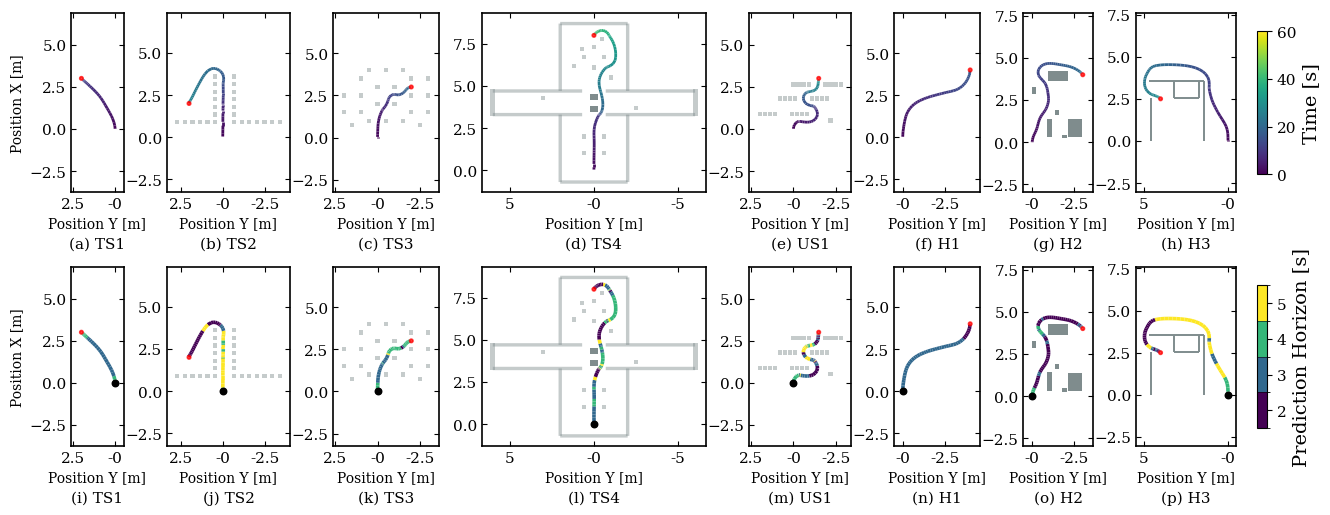}
    \caption{Executed trajectories of D3DWA. Top row (a)–(h): coloured by elapsed time. Bottom row (i)–(p): the same trajectories coloured by the prediction horizon $\tau$ selected by the agent. The robot starts at the origin and the goal is marked in red.}
    \label{fig:trajtau}
\end{figure*}

\subsection{Simulation Environments}
\label{sec:sim-env}

The simulation environments are shown in Fig.~\ref{fig:scenes}. Two groups of
scenarios were considered. The first group consists of four environments used
during training, denoted as \texttt{TS1--TS4}. The second group evaluates
generalization to environments not used for training. \texttt{US1} retains a
similar overall layout but contains narrower passages, whereas
\texttt{H1--H3} are unseen hospital environments with different layouts.

The robot started from $(0.0,0.0)$ in all environments. Goal positions were
fixed for evaluation, and all obstacles remained static throughout each trial.
These environments include both relatively open spaces and narrow or cluttered
passages, allowing the effect of adapting the prediction horizon to be evaluated
under different local navigation conditions.

\subsection{Simulation Results}
\label{sec:sim-results}

Table~\ref{tab:main} summarizes the simulation results. The three fixed DWA
configurations show strong sensitivity to the selected evaluation weights.
DWA~I, which emphasizes speed, reaches only two of the eight goals, whereas
DWA~II reaches five. DWA~III is the strongest fixed-parameter baseline,
completing six of the eight environments. However, no fixed parameter setting
successfully completes all environments, indicating that a single set of DWA
parameters is insufficient across environments with different levels of
congestion.

DQDWA~\cite{10237217} completes five of the eight environments. Replacing its
tabular state representation with the continuous observation used by D3DWA,
while keeping the prediction horizon fixed, improves performance:
D3DWA (w/o horizon) completes six environments. In the unseen environments,
it reduces the navigation time from $46.09$\,s to $37.39$\,s in
\texttt{US1} and from $58.41$\,s to $36.45$\,s in \texttt{H1}, and also
successfully completes \texttt{H3}, where DQDWA times out. These results
suggest that the continuous-state D3QN formulation improves adaptation
compared with the tabular DQDWA baseline. Because this comparison changes
both the state representation and the learning algorithm, the improvement is
not attributed to either factor in isolation.

The effect of prediction-horizon adaptation is most clearly observed by
comparing D3DWA with D3DWA (w/o horizon). The weights-only variant times out
in both \texttt{TS2} and \texttt{TS4}, whereas the full D3DWA successfully
reaches the goal in both environments. Consequently, D3DWA is the only method
that completes all eight environments. It also achieves the shortest
navigation time in six of the eight environments.

These results indicate that adapting the evaluation weights alone improves
navigation performance but does not fully address changes in local free space.
Joint adaptation of the evaluation weights and prediction horizon expands the
range of environments that can be successfully navigated while retaining the
underlying DWA planner.
The trajectory examples in
Figs.~\ref{fig:traj-dwa3}--\ref{fig:trajtau} further illustrate the differences
in navigation behavior among the fixed-parameter, tabular, and proposed
methods.

The bottom row of Fig.~\ref{fig:trajtau} shows the same trajectories coloured by the prediction horizon $\tau$ selected at each control step. The horizon is not constant within a run, and its variation is spatially structured: long horizons ($\tau = 5\,\mathrm{s}$) occur along the open straight approach in \texttt{TS2} and the corridor sweep in \texttt{H3}, whereas short horizons ($\tau = 2$--$3\,\mathrm{s}$) dominate the cluttered and constricted regions of \texttt{TS3}, \texttt{TS4}, \texttt{US1}, and \texttt{H2}. In \texttt{H1}, where no obstacle lies along the executed path, the selected horizon stays fairly constant. \texttt{TS2} shows both regimes within a single route, switching from $\tau = 5\,\mathrm{s}$ on the straight segment to $\tau = 2\,\mathrm{s}$ at the turn into the narrowed goal region; D3DWA (w/o horizon), which holds $\tau = 4\,\mathrm{s}$ throughout, times out there. The horizon therefore provides a degree of freedom the weights cannot substitute for.

\begin{figure}[t]
  \centering
  \includegraphics[width=\columnwidth]{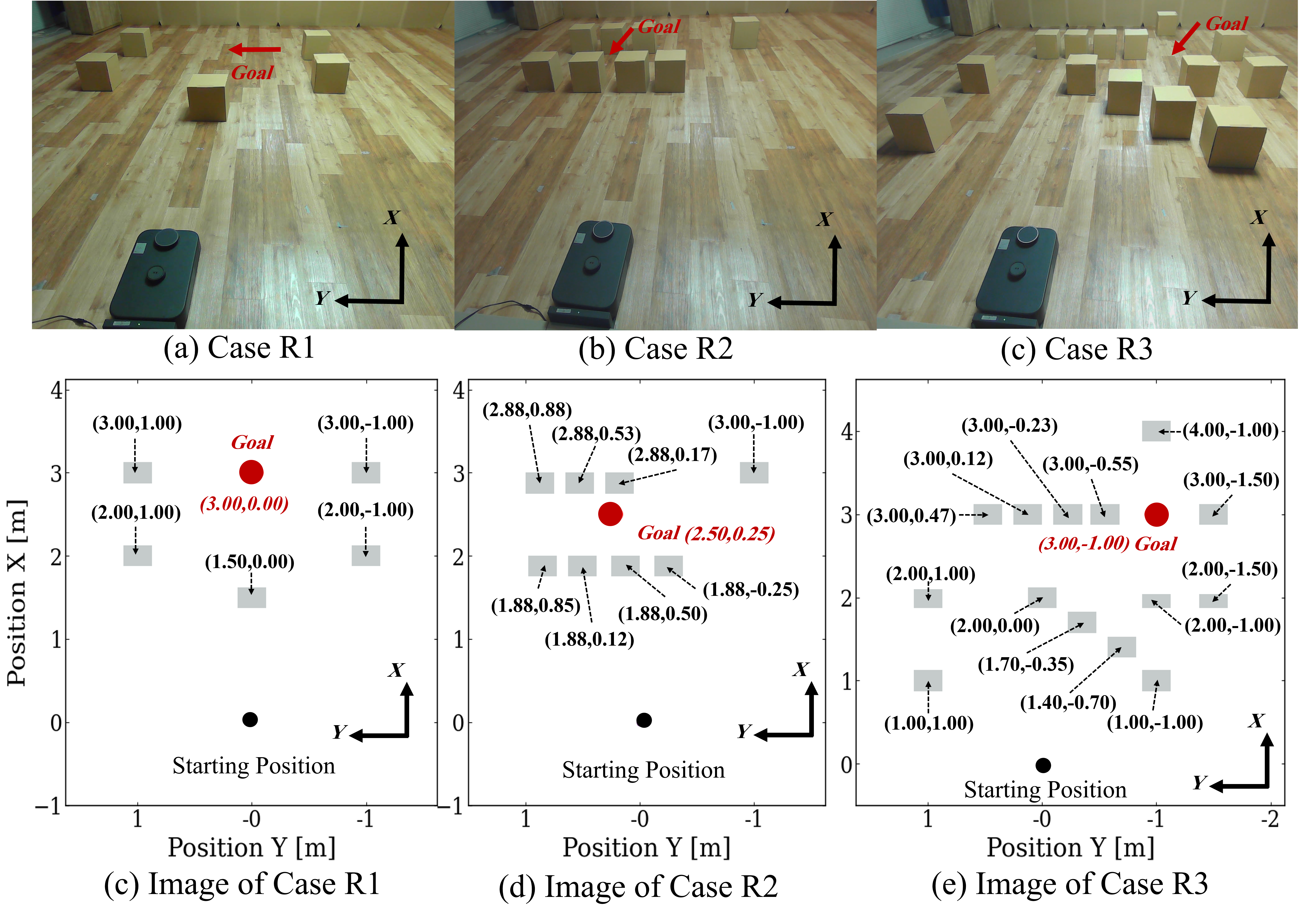}
  \caption{\textbf{Real-robot experimental setup.}}
  \label{fig:real}
\end{figure}

\section{Real World Experiments}
\subsection{Experimental Setup}

\begin{figure*}[t]
  \centering
  \includegraphics[width=\textwidth]{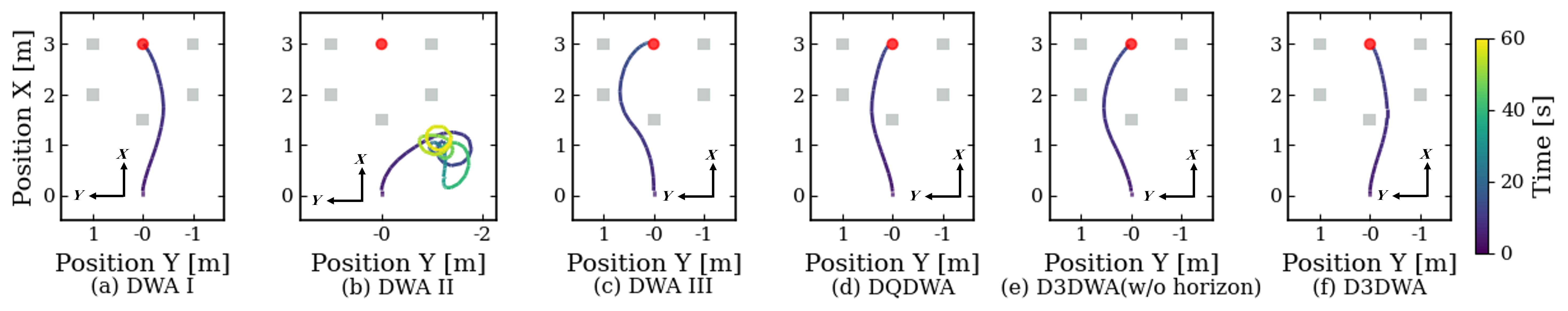}
  \caption{\textbf{Case R1 Trajectories.}}
  \label{fig:case1Traj}
\end{figure*}

\begin{figure*}[t]
  \centering
  \includegraphics[width=\textwidth]{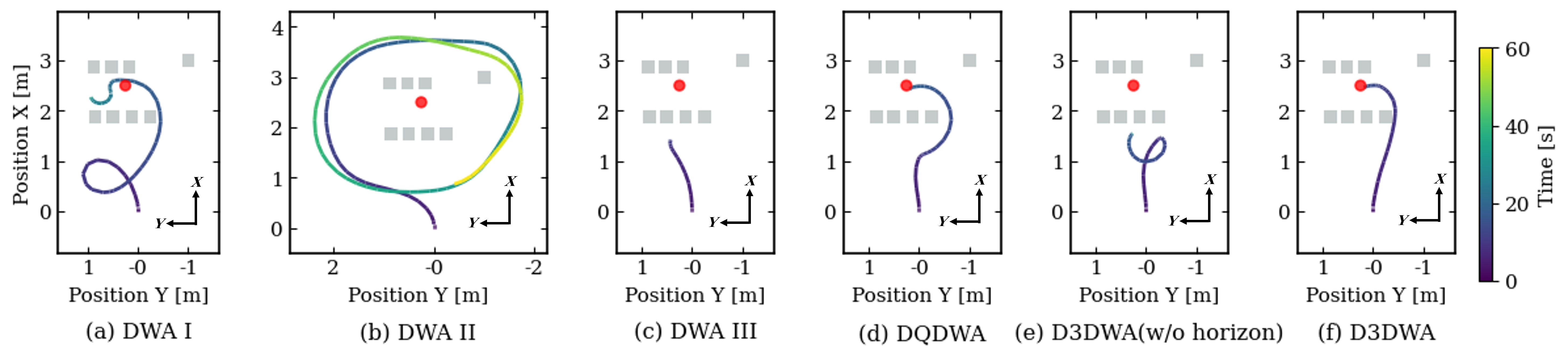}
  \caption{\textbf{Case R2 Trajectories.}}
  \label{fig:case2Traj}
\end{figure*}

\begin{figure*}[t]
  \centering
  \includegraphics[width=\textwidth]{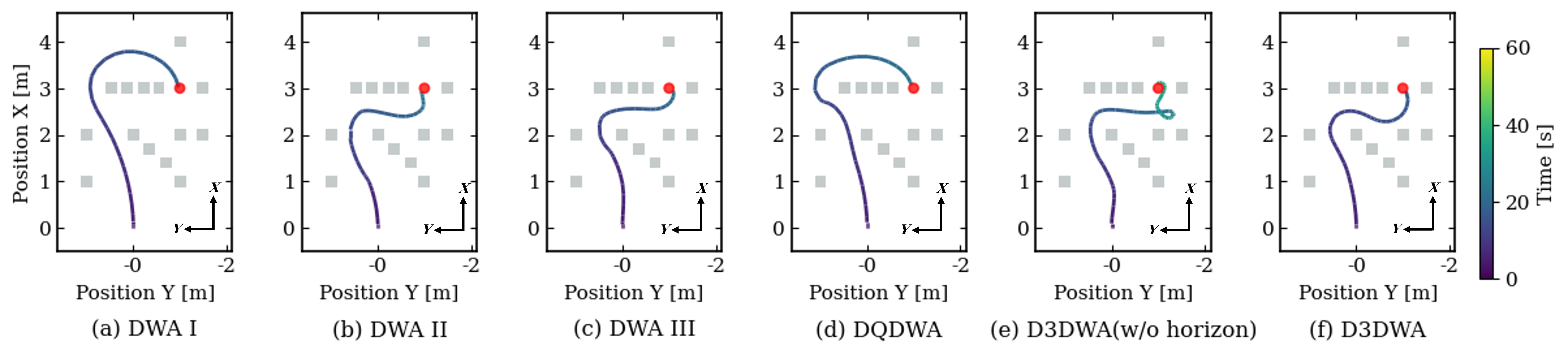}
  \caption{\textbf{Case R3 Trajectories.}}
  \label{fig:case3Traj}
\end{figure*}

The proposed method was evaluated on a physical Kachaka mobile robot in the
same room used to construct the simulation environments, as shown in
Fig.~\ref{fig:real}. Three static-obstacle
configurations, denoted as Cases~R1--R3, were used to evaluate navigation
performance under different spatial arrangements of obstacles. The obstacle
layouts and executed trajectories for the three configurations are shown in
Figs.~\ref{fig:case1Traj}--\ref{fig:case3Traj}.

Case~R1 places five obstacles between the start position and a goal at
$(3.0,0.0)$. Case~R2 places eight obstacles in two staggered rows before a
goal at $(2.5,0.25)$, producing narrower passages and a more constrained
navigation problem than Case~R1. Case~R3 places fourteen obstacles, the most of
the three, with the goal off the initial heading at $(3.0,-1.0)$. Its density
is higher than in Case~R2 but its passages are wider, so it probes efficiency
in clutter rather than goal-reaching in a constriction.

The policies learned in simulation were transferred directly to the physical
robot without retraining or fine-tuning. 
The same six methods and the same three metrics as in the simulation were
used, with one trial per method and experimental condition.

\begin{figure}[t]
  \centering
  \includegraphics[width=\columnwidth]{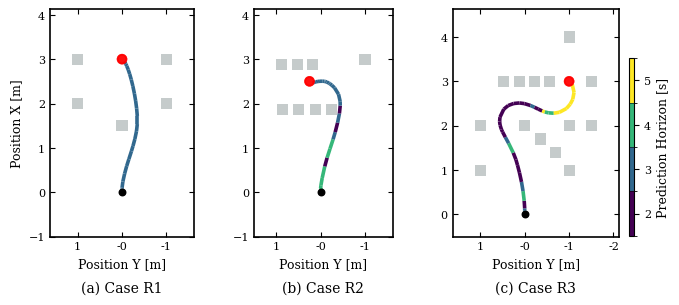}
  \caption{\textbf{D3DWA in all cases coloured by prediction horizon $\tau$ selected by the agent.}}
  \label{fig:realprediction}
\end{figure}

\subsection{Experimental Results}

Table~\ref{tab:real} summarizes the real-robot results, while
Figs.~\ref{fig:case1Traj}--\ref{fig:case3Traj} show the corresponding
trajectories. The results demonstrate different characteristics of the three
experimental configurations.

In Case~R1, five of the six methods reached the goal, with only DWA~II timing
out. DWA~I achieved the shortest navigation time of $12.45$\,s, whereas D3DWA
produced the shortest trajectory ($3.04$\,m) and the lowest posture displacement
($1.20$\,rad). These results indicate that, in the relatively less constrained
configuration, several parameter settings remain feasible, and the primary
difference among the successful methods is navigation efficiency rather than
goal-reaching capability.

\begin{table}[t]
\centering
\caption{Real-robot results. One trial per condition.
\textsf{T} denotes timeout. Best values for each case and metric are shown in bold.}
\label{tab:real}

\scriptsize
\setlength{\tabcolsep}{2.0pt}
\renewcommand{\arraystretch}{1.10}

\resizebox{\columnwidth}{!}{%
\begin{tabular}{l*{3}{ccc}}
\toprule
\multirow{2}{*}{Method}
& \multicolumn{3}{c}{Case R1}
& \multicolumn{3}{c}{Case R2}
& \multicolumn{3}{c}{Case R3} \\
\cmidrule(lr){2-4}
\cmidrule(lr){5-7}
\cmidrule(lr){8-10}

& Time & TL & PD
& Time & TL & PD
& Time & TL & PD \\
\midrule

DWA I $(1,1,2)$
& \textbf{12.45} & 3.14 & 1.35
& \textsf{T} & -- & --
& 19.38 & 5.89 & \textbf{4.07} \\

DWA II $(1,2,1)$
& \textsf{T} & -- & --
& \textsf{T} & -- & --
& 24.76 & 4.45 & 5.07 \\

DWA III $(2,1,1)$
& 19.10 & 3.43 & 2.93
& \textsf{T} & -- & --
& 22.62 & \textbf{4.33} & 5.19 \\

DQDWA~\cite{10237217}
& 13.99 & 3.08 & 1.41
& 20.22 & 3.36 & 5.10
& 24.27 & 6.02 & 6.18 \\

D3DWA (w/o horizon)
& 14.91 & 3.22 & 1.72
& \textsf{T} & -- & --
& 38.70 & 5.51 & 11.64 \\

\textbf{D3DWA (proposed)}
& 13.70 & \textbf{3.04} & \textbf{1.20}
& \textbf{15.70} & \textbf{2.98} & \textbf{2.77}
& \textbf{17.52} & 4.61 & 5.52 \\

\bottomrule
\end{tabular}%
}
\end{table}
Case~R2 provides a more demanding test. Among the six methods, only DQDWA and
D3DWA successfully reached the goal. All three fixed-parameter DWA variants and
D3DWA (w/o horizon) timed out. D3DWA completed the task in $15.70$\,s, compared
with $20.22$\,s for DQDWA, while also reducing the trajectory length from
$3.36$\,m to $2.98$\,m and the posture displacement from $5.10$\,rad to
$2.77$\,rad. The failure of D3DWA (w/o horizon), despite using the same
continuous observation and D3QN formulation, is particularly informative:
adapting the evaluation weights alone was insufficient to navigate this
configuration, whereas jointly adapting the prediction horizon enabled the
robot to complete the task.

Case~R3 provides a complementary evaluation in which all six methods
successfully reached the goal. Even under this condition, D3DWA achieved the
shortest navigation time of $17.52$\,s. This corresponds to a reduction of
$27.8\%$ relative to DQDWA ($24.27$\,s) and $54.7\%$ relative to D3DWA
(w/o horizon) ($38.70$\,s). D3DWA also followed a substantially shorter
trajectory than DQDWA ($4.61$\,m versus $6.02$\,m) and required less posture
change ($5.52$\,rad versus $6.18$\,rad). The shortest trajectory in this case
was obtained by DWA~III ($4.33$\,m), while DWA~I achieved the lowest posture
displacement ($4.07$\,rad). Thus, although D3DWA does not minimize every
individual metric in Case~R3, it achieves the highest navigation efficiency
in terms of completion time while maintaining competitive trajectory length
and posture displacement.

Overall, D3DWA and DQDWA were the only methods that completed all three
real-robot configurations, and D3DWA was faster in Cases~R2 and R3. The
weights-only variant is the informative case: it succeeded in Cases~R1 and R3
but timed out in Case~R2, supporting the simulation finding that horizon
adaptation matters when the locally available free space changes along the
route.

Fig.~\ref{fig:realprediction} shows the real-robot D3DWA trajectories coloured by the
selected prediction horizon. The policy holds an almost constant
$\tau = 3\,\mathrm{s}$ in the unconstrained Case~R1, stays at the short
end of the candidate set in Case~R2 while working through the staggered rows,
and in Case~R3 switches from $\tau = 2\,\mathrm{s}$ in the cluttered lower
region to $\tau = 5\,\mathrm{s}$ on the open final approach; showing the complementary effect on efficiency, where the weights-only variant reached the goal but required $38.70\,\mathrm{s}$ against $17.52\,\mathrm{s}$ for D3DWA.

The discrepancy between simulation and the physical experiments can be
attributed in part to real LiDAR noise, wheel slip, and odometry drift.
LiDAR measurements may temporarily underestimate obstacle distances, causing
candidate trajectories that are marginally feasible in simulation to be
rejected on the physical robot. With a fixed $4$\,s prediction horizon, this
effect can substantially reduce the set of admissible trajectories in narrow
spaces. D3DWA can instead select a shorter prediction horizon according to the
current observation, allowing the underlying DWA planner to retain feasible
local motions without changing its trajectory-generation or collision-checking
mechanism.

\section{Conclusion}
This paper presented D3DWA, an adaptive local-planning framework that uses a Dueling Double Deep Q-Network to jointly select the DWA evaluation weights and prediction horizon at every control step, while preserving the original DWA trajectory-generation and collision-checking mechanisms. In eight simulated environments, including unseen layouts, D3DWA was the only method to reach every goal and achieved the shortest navigation time in six environments. The weight-only ablation failed in environments where the full D3DWA succeeded, indicating that adapting the prediction horizon provides an additional capability that cannot be achieved by adjusting the evaluation weights alone. Real-robot experiments further supported this finding: D3DWA completed all three tested configurations, including the constrained Case~R2 in which the weight-only variant timed out. These results demonstrate that jointly adapting the evaluation weights and prediction horizon can improve the robustness and efficiency of DWA across varying local navigation conditions.
Since the current real-robot evaluation is limited to a single robot platform, future work will investigate the generalizability of the proposed method across multiple robot platforms and environments.


\vfill

\end{document}